%% file: main.tex
\documentclass[letterpaper, 10 pt, conference]{ieeeconf}  
\IEEEoverridecommandlockouts                              
\usepackage[english]{babel}
\usepackage[T1]{fontenc}

\usepackage[nolist]{acronym}
\usepackage[cmex10]{amsmath}
\usepackage{amssymb}
\usepackage{bm}
\usepackage{booktabs}
\usepackage{color}
\usepackage{float}
\usepackage{graphicx}
\usepackage{hyperref}
\usepackage{import}
\usepackage{mathtools}
\usepackage{multirow}
\usepackage{outline}
\usepackage{paralist}
\usepackage{siunitx}
\usepackage{stmaryrd}
\usepackage{systeme}
\usepackage{units}
\usepackage{url}
\usepackage{tikz}
\usepackage{booktabs}
\usepackage{amsfonts}
\usepackage{amssymb}
\usetikzlibrary{tikzmark}
\usepackage{physics}
\usepackage{subcaption}
\usepackage{float}
\usepackage{balance}


\title{\LARGE \bf Autonomous Navigation for Quadrupedal Robots with Optimized Jumping through Constrained Obstacles
\thanks{* Authors have contributed equally.}
\thanks{$^1$ Experimental videos can be found at \url{https://youtu.be/5pzJ8U7YyGc}.}
\thanks{All authors are affiliated with UC Berkeley, USA. \tt\small\{scott\_gilroy, delau, lzyang, eizaguir, kbiermayer, xax, mengtisun, ayush.agrawal, zengjunsjtu, zhongyu\_li, koushils\}@berkeley.edu}
}
\author{Scott Gilroy*, Derek Lau*, Lizhi Yang*, Ed Izaguirre, Kristen Biermayer, Anxing Xiao, Mengti Sun, \\Ayush Agrawal, Jun Zeng, Zhongyu Li, Koushil Sreenath}

\begin{document}
\maketitle

\begin{abstract}
Quadrupeds are strong candidates for navigating challenging environments because of their agile and dynamic designs. 
This paper presents a methodology that extends the range of exploration for quadrupedal robots by creating an end-to-end navigation framework that exploits walking and jumping modes.
To obtain a dynamic jumping maneuver while avoiding obstacles, dynamically-feasible trajectories are optimized offline through collocation-based optimization where safety constraints are imposed. 
Such optimization schematic allows the robot to jump through window-shaped obstacles by considering both obstacles in the air and on the ground. 
The resulted jumping mode is utilized in an autonomous navigation pipeline that leverages a search-based global planner and a local planner to enable the robot to reach the goal location by walking.
A state machine together with a decision making strategy allows the system to switch behaviors between walking around obstacles or jumping through them.
The proposed framework is experimentally deployed and validated on a quadrupedal robot, a Mini Cheetah, to enable the robot to autonomously navigate through an environment while avoiding obstacles and jumping over a maximum height of 13 cm to pass through a window-shaped opening in order to reach its goal. (Video$^1$) 
\end{abstract}

\input{sections/introduction.tex}
\input{sections/framework.tex}
\input{sections/modeling.tex}

\input{sections/trajectory-generation.tex}

\input{sections/planners.tex}

\input{sections/controller.tex}
\input{sections/results.tex}
\input{sections/conclusion.tex}

\section*{Acknowledgments}
This work is supported in part by National Science Foundation Grants CMMI-1944722. The authors would like to thank Professor Sangbae Kim, the MIT Biomimetic Robotics Lab, and NAVER LABS for providing the Mini Cheetah simulation software and lending the Mini Cheetah for experiments.

\balance
{
\bibliographystyle{IEEEtran}
\bibliography{bib/bibliography}
}

\begin{acronym}
\acro{HP}{high-pass}
\acro{LP}{low-pass}
\end{acronym}

\end{document}

%% file: sections/introduction.tex
\section{Introduction}
\label{sec:Introduction}
\begin{figure}
  \centering
  \includegraphics[width=\linewidth]{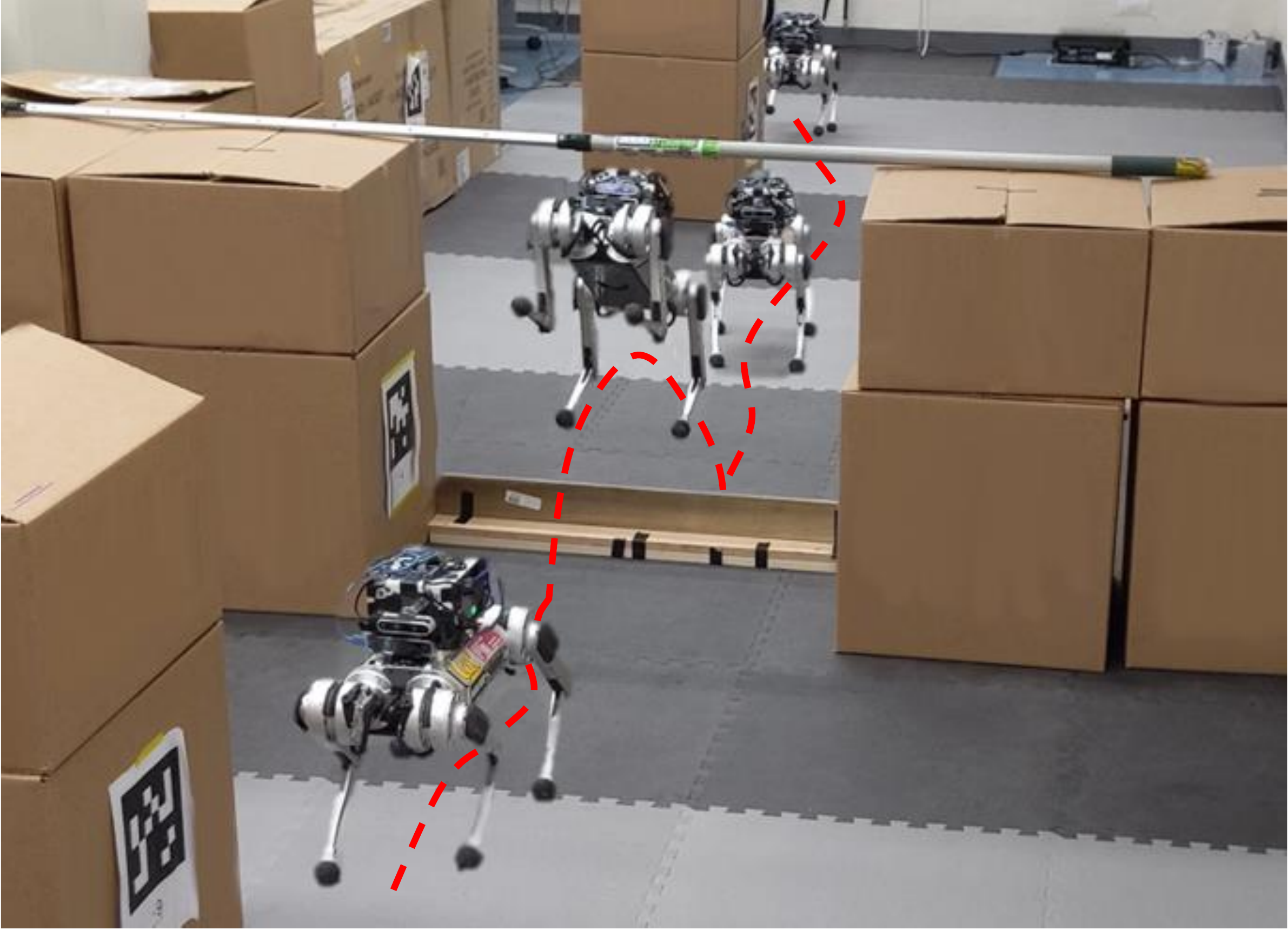}
  \caption{A Mini Cheetah travelling in a congested environment autonomously by navigating around obstacles and jumping through an obstacle in order to reach the goal location.}
  \label{fig:cover}
\end{figure}

Quadrupedal robots have a wide range of mobility modes that allow for greater applicability in navigating challenging environments.
Unlike wheeled robots that have difficulty in traversing uneven terrain, quadrupeds display the ability to adapt to rough terrain~\cite{mastalli2020motion, yang2020dynamic}, and achieve dynamic motions, such as bounding over obstacles~\cite{park2015online}, depicting potentials of usability in unstructured environments and of being dynamic service robots~\cite{xiao2021robotic}.
Recently, research has shown that quadrupeds are capable of performing a wide range of jumps~\cite{katz2019mini, nguyen2019optimized}, allowing these robots to reach areas that were previously inaccessible due to the discrete change of ground level. 
While prior work has demonstrated the feasibility of jumping maneuvers on quadrupedal robots~\cite{kim2020vision}, the additional constraints of jumping through constrained obstacles, \textit{i.e.}, jumping while avoiding a collision, was not considered.
Moreover, autonomous navigation exploiting the jumping ability to avoid obstacles of these robots in congested environments has not been demonstrated.   
In this paper, we focus on tackling a challenging scenario where the traversable region is narrow and window-shaped, \textit{i.e.}, there are obstacles on the ground, the sides and in the air, as shown in Fig.~\ref{fig:cover}. 
By solving such a problem, we seek to develop a navigation framework for quadrupedal robots with multi-modal transitions that include walking modes to move around obstacles and jumping maneuvers to hop over obstacles in order to reach a specified goal location.

\begin{figure*}[!htp]
  \centering
  \includegraphics[width=0.9\textwidth]{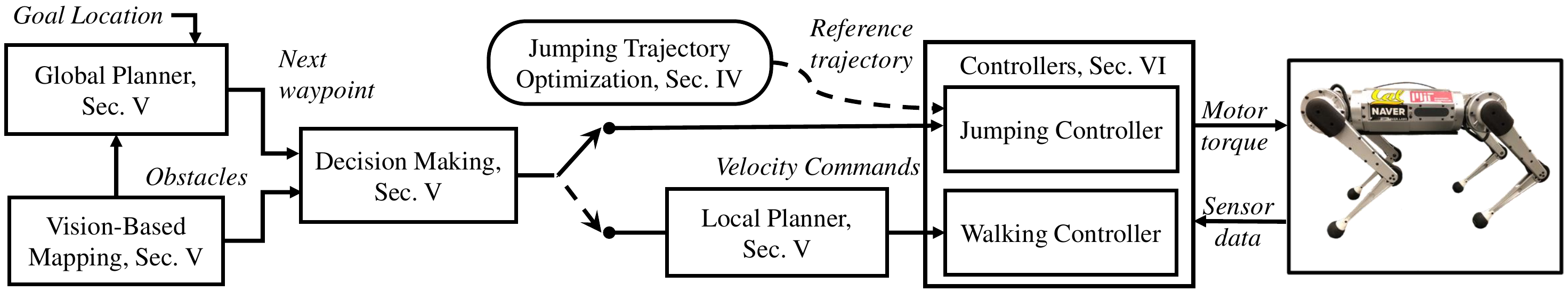}
  \caption{Proposed framework for navigating in confined spaces with an optimized jumping mode. This autonomous pipeline allows the robot to sense the environment with an RGB-Depth camera and a tracking camera. A global path to reach the goal location is obtained by a search-based planner. A state machine determines robot locomotion modes, such as walking and jumping, in order to move around or jump over obstacles that are not traversable by walking.}
  \label{fig:framework}
\end{figure*}

\subsection{Related Work}

Prior work has demonstrated that quadrupeds have a promising ability to move freely through an environment. %
One approach for motion planning utilizes inverse kinematics, where collision-free trajectories are generated by solving for body poses and foot contact positions~\cite{geisert2019contact, buchanan2021perceptive}.
While this method was presented to solve for trajectories quickly, the use of inverse kinematics cannot prevent singularities from occurring.  Singularities could easily present when moving legs to avoid height-constrained environments.
Alternatively, jumps can be accomplished by scaling the impulse applied during bounding to complete running jumps as seen in~\cite{park2014quadruped}.  However, this work fails to consider window-shaped obstacles. 
Trajectory optimization has been utilized to generate jumps in \cite{park2015online} but lacks the integration with navigating through unknown environments.

Motion planning for agile movements such as jumping can also be formulated as an optimization problem, where the dynamics of the robot as well as obstacles are incorporated in the constraints to obtain optimal state trajectories~\cite{winkler2018gait, neunert2017trajectory, katz2019mini}. Robots such as SALTO-1P~\cite{yim2020precision}, the MIT Cheetah~\cite{park2014quadruped, park2015online}, and ANYmal~\cite{hutter2017anymal} have all displayed capabilities in performing jumps and navigating around the environment. A common approach in such techniques is to use a reduced order model of the robot by binding the left and right legs to improve computation speed while not sacrificing the ability to jump along the sagittal plane~\cite{nguyen2019optimized, kim2020vision}. Various control techniques such as Model Predictive Control (MPC)~\cite{di2018dynamic}, Whole-Body Impulse Control (WBIC)~\cite{yang2020dynamic, kim2019highly} or PD control \cite{nguyen2019optimized, kim2020vision} can then be used to track these reference trajectories. These works, however, do not consider jumping in highly constrained environments such as jumping through window-shaped obstacles.

Our approach leverages recent results in trajectory optimization using a duality-based approach~\cite{zhang2020optimization} which allows us to convert obstacle avoidance criteria between two convex regions, \textit{i.e.}, maintaining a minimum distance, into a differentiable and smooth non-convex avoidance constraint.
The smoothness property allows the use of general-purpose gradient- and Hessian-based optimization algorithms in trajectory generation.
This approach has been applied to different robots, such as quadrotors~\cite{zeng2020differential} and wheeled robots~\cite{zhang2018autonomous}.
This motivates us to apply this formulation of obstacle avoidance to the legged systems, which could be exploited to generate a collision-free trajectory through height-constraint obstacles.

\subsection{Contributions}
The main contributions of this paper are the development of one of the first end-to-end navigation autonomy for quadrupedal robots to travel in an unstructured environment through multi-modal transition. 
Such framework leverages two locomotion modes, planar walking and forward jumps. 
In order to achieve dynamic jumping maneuvers while avoiding obstacles, a multi-phased collocation based nonlinear optimization is firstly developed to generate a trajectory of the robot to jump through window-shaped obstacles while respecting the robot's physical limitations. 
A controller is also introduced to enable the robot to achieve the optimized jump online. 
By integrating a state machine, path planners, and the proposed jumping controller into the navigation pipeline, we firstly enable a quadrupedal robot, the Mini Cheetah, to intelligently choose either walking around obstacles or jumping over obstacles without collision while traveling in environments that are previously nontraversable by only the walking mode.

%% file: sections/framework.tex
\section{Framework}
\label{sec:framework}
The proposed navigation autonomy using quadrupedal robots developed in this paper is illustrated in Fig.~\ref{fig:framework}.
The environment is perceived by an RGB-Depth camera, and obstacles are registered into maps which include the maximum obstacle height on each grid.
After being given a target goal location, a global planner using A*~\cite{hart1968formal} is constructed to find a sequence of collision-free waypoints on the map from the robot's current position to the goal location. 
A state machine is designed to switch between the robot locomotion modes of walking and jumping. 
A local planner is later developed to obtain desired walking velocity profiles for the robot to track a waypoint selected from the global path. 
If the planned waypoints cross a non-traversable obstacle which the robot is capable of jumping over, the state machine will prepare the robot to jump. 
Mapping, planners, and the state machine use Visual Intertial Odometry~(VIO) to estimate the robot's current state and they are introduced in Sec.~\ref{sec:planners}.
In order to enable the robot to attain a dynamic jumping mode while considering the surrounding confined space, such as a window-shaped obstacle, trajectory optimization is formulated offline to obtain dynamically-feasible joint profiles for the robot while imposing collision-avoidance constraints. 
This optimization is discussed in Sec.~\ref{sec:trajectory-generation} and it is based on the robot planar model presented in Sec.~\ref{sec:modelling}.
A jumping controller developed in Sec.~\ref{sec:control} empowers the robot to perform the optimized trajectory, resulting in a safe landing on the target point behind the obstacles.
Such model-based vision-to-torque navigation pipeline is deployed on a Mini Cheetah and is validated in experiments presented in Sec.~\ref{sec:results}.
Conclusion and future work is discussed in Sec.~\ref{sec:conclusion}.




%% file: sections/modeling.tex
\section{Model and Dynamics}
\label{sec:modelling}
In this section, we characterize the quadrupedal jumping behavior with different phases.
According to the jumping behavior, the robot is simplified under a planar model and the corresponding dynamics is also introduced.

\subsection{Simplified Model}
\begin{figure}[!]
    \centering
    \includegraphics[width=0.75\linewidth]{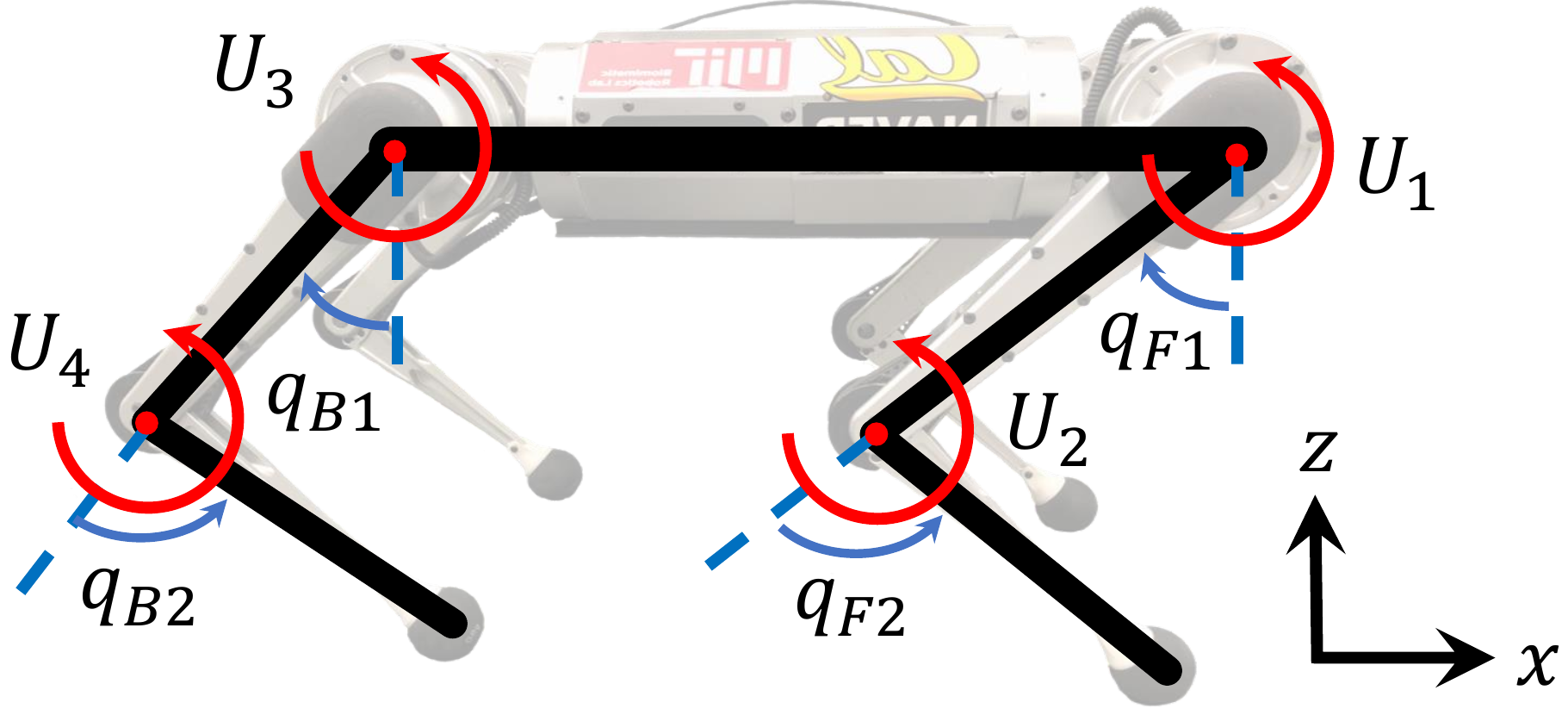}
    \caption{Simplified planar model used for quadruped jumping. The planar model consists of base position, orientation together with four joints. The inputs of this planar model are four torques acting on joints.
    }
    \label{fig:planar-model}
\end{figure}

The jumping behavior of the quadruped motivates us to propose a simplified model that takes advantage of the robot's symmetry. 
A planar model with two legs is used to capture the jumping behaviors for the quadruped and is shown in Fig. \ref{fig:planar-model}. 
The system states $\mathbf{x}$ and control inputs $\mathbf{u}$ of this simplified model are as follow,
\begin{align}
\mathbf{q} &:= [q_x, q_z, q_{\theta}, q_{F1}, q_{F2}, q_{B1}, q_{B2}]^T \in \mathbb{R}^7 \\
\mathbf{x} &:= [\mathbf{q}; \dot{\mathbf{q}}] \in \mathbb{R}^{14} \\
\mathbf{u} &:= [\tau_{F1}, \tau_{F2}, \tau_{B1}, \tau_{B2}]^T \in \mathbb{R}^4,
\end{align}
where $q_x$, $q_z$ and $ q_{\theta}$ represent the robot's base position and pitch angle, respectively. 
$\mathbf{q} = [q_{F1}, q_{F2}, q_{B1}, q_{B2}]^T$ is the hip and knee joint angles of the front and back leg of the simplified model and $\mathbf{u}$ represents the input of this planar model which consists of torques acting on the front and back leg. 
The simplification from the original four-legged model~\cite{bosworth2015super} allows us to reduce the computational complexity by reducing from the 18 Degrees-of-Freedom (DOF) model for the 3D quadruped into 7 DOF for the planar system.

We also denote the robot's pose in 3D by using $q_y$ to represent the robot's position in the lateral direction and $q_\phi$ to represent the robot's turning yaw.

\subsection{Quadrupedal Jumping}
\label{subsec:jump-phases}
\begin{figure}[!]
    \centering
    \includegraphics[width=\linewidth]{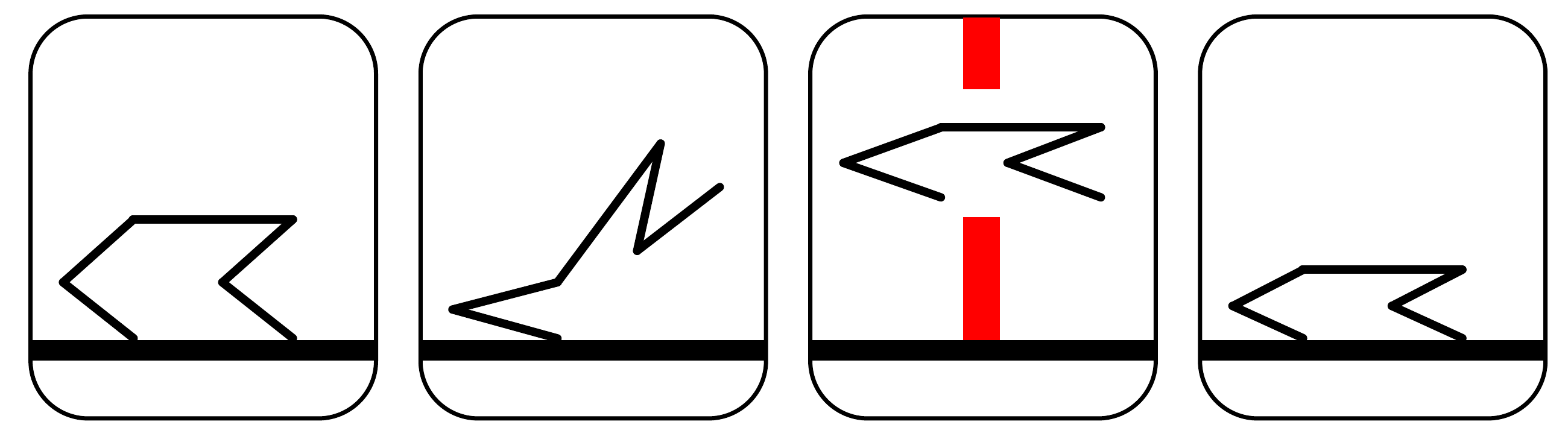}
    \caption{Jumping behavior of a simplified quadrupedal robot. The phases can be distinguished as all feet contact, rear feet contact, flight, and landing, respectively. The possible window-shaped obstacles are colored in red in the flight phase.}
    \label{fig:jumping-behavior}
\end{figure}

The jumping maneuver of quadrupedal robots is featured as four different phases as illustrated in Fig.~\ref{fig:jumping-behavior}, which include: all feet contact, rear feet contact, flight, and landing:
\begin{enumerate}
    \item \textit{All feet contact phase}: Both the front and back feet are in direct contact with the ground. To transition to the rear feet contact phase, the quadruped pushes off the ground using the front legs and rotates upon the back feet.
    \item \textit{Rear feet contact phase}: The back legs extend allowing for the quadruped to push off the ground. When the contact forces of the back legs become zero, the robot switches into the flight phase.
    \item \textit{Flight phase}: The whole robot is under free fall while being able to reconfigure in the air allowing itself to perform different tasks, such as avoiding obstacles.
    \item \textit{Landing phase}: At the end of the flight phase, the robot remakes contact with the ground on all feet.
\end{enumerate}

\subsection{Simplified Dynamics}
The dynamics of the planar model can be modeled based upon the Lagrangian equations of motion~\cite{westervelt2018feedback}:
\begin{equation}
\label{eq:dynamicsEqn}
    \mathbf{D}(\mathbf{q})\ddot{\mathbf{q}} + \mathbf{C}(\mathbf{q}, \dot{\mathbf{q}})\dot{\mathbf{q}} + \mathbf{G}(\mathbf{q}) = \mathbf{B}\mathbf{u} + \mathbf{J}(\mathbf{q})^{T} \mathbf{T},
\end{equation}
where $\mathbf{D}(\mathbf{q})$ is the Mass-Inertia matrix, $\mathbf{C}(\mathbf{q}, \dot{\mathbf{q}})$ is the Coriolis matrix, $\mathbf{G}(\mathbf{q})$ is the gravity vector, $\mathbf{B}$ is the input mapping matrix, $\mathbf{u}$ is the joint torques and $\mathbf{T} := [T_{Fx}, T_{Fz}, T_{Bx}, T_{Bz}]^T \in \mathbb{R}^4$ denotes the contact forces acting on the front and back leg from the ground. Additionally, the following constraint is needed for the feet to be in contact with the ground.
\begin{equation}
    \label{eq:groundConstraintEqn}
    \mathbf{J}^{(i)}(\mathbf{q})\ddot{\mathbf{q}} + \mathbf{\dot{J}}^{(i)}(\mathbf{q})\dot{\mathbf{q}} = 0.
\end{equation}
By combining equations \eqref{eq:dynamicsEqn} and \eqref{eq:groundConstraintEqn}, we could capture the system dynamics at $i$-th phase with notations as follow,
\begin{equation}
    \dot{\mathbf{x}} = f^{(i)} (\mathbf{x}, \mathbf{u}, \mathbf{T})
\end{equation}

%% file: sections/trajectory-generation.tex
\section{Jumping Trajectory Generation}
\label{sec:trajectory-generation}
After having introduced the simplified model and dynamics for quadrupedal jumping, we are going to illustrate the trajectory generation for jumping mode using a collocation-based optimization.
The optimization formulation is introduced in Sec. \ref{subsec:formulation}.
After that, cost function design, physical constraints, and safety constraints are presented in Sec. \ref{subsec:cost-function}, \ref{subsec:physical-constraints}, and \ref{subsec:obstacle-avoidance}, respectively.

\subsection{Formulation}
\label{subsec:formulation}
We propose an optimization-based trajectory generation for the jumping motion, which is stated as follows,
\begin{subequations}
\label{eq:optimization-formulation}
    \begin{align}
        \min_{\mathbf{x}, \mathbf{\dot{x}}, \mathbf{u}, \mathbf{T}} \ & J(\mathbf{x},\dot{\mathbf{x}}, \mathbf{u}, \mathbf{T}) \label{subeq:opti} \\
        \text{s.t.} \ & \mathbf{x}(t_{k+1}) = \mathbf{x}(t_k) \ + \frac{\Delta t^{(i)}}{2} (\mathbf{\dot{x}}(t_{k+1}) + \mathbf{\dot{x}}(t_k)) \label{subeq:collocation} \\
        & \dot{\mathbf{x}}(t_{k+1}) = f^{(i)}(\mathbf{x}(t_{k}), \mathbf{u}(t_{k}), \mathbf{T}(t_k)) \label{subeq:dynamics} \\
        & \mathbf{x}(t_0) = \mathbf{x}_0 \label{subeq:initial-condition}
    \end{align}
\end{subequations}
where $\Delta t^{(i)} = T_i / N_i$ for the $i$-th phase (as described in Sec.~\ref{subsec:jump-phases}) with the duration of the phase being $T_i$ and the number of collocation nodes in the phase being $N_i$.
The optimization \eqref{eq:optimization-formulation} optimizes of the cost function $J(\mathbf{x},\dot{\mathbf{x}}, \mathbf{u}, \mathbf{T})$ over $N = \sum_{i = 1}^{4} N_i$ collocation nodes.
The constraint on the initial condition $\mathbf{x}_0$  is in \eqref{subeq:initial-condition}.
The trapezoidal constraints between nodes are in \eqref{subeq:collocation} and system dynamics are considered as constraints in \eqref{subeq:dynamics}.


\subsection{Cost Function}
\label{subsec:cost-function}
The cost function $J(\mathbf{x},\dot{\mathbf{x}}, \mathbf{u}, \mathbf{T})$ is formulated as follow,
\begin{equation}
\label{eq:cost-function-design}
    \begin{split}
        J =& (\mathbf{q}(t_{N+1})-\mathbf{q}_0)^T P_f (\mathbf{q}(t_{N+1})-\mathbf{q}_0) \\
        & + \sum_{k=0}^{N+1} (\dot{\mathbf{q}}^T(t_k) Q_{\dot{\mathbf{q}}} \dot{\mathbf{q}}(t_k) + \ddot{\mathbf{q}}^T(t_k) Q_{\ddot{\mathbf{q}}} \ddot{\mathbf{q}}(t_k) \\
        & + \mathbf{T}(t_k)Q_{\mathbf{T}}\mathbf{T}(t_k) + \mathbf{u}^T(t_k) R_{\mathbf{u}} \mathbf{u}(t_k))\\
        &  + \sum_{i=1}^{2} P_i T_i + \sum_{i=1}^{N_2} P_{\delta} \delta
    \end{split}
\end{equation}
where $Q_{\mathbf{\dot{q}}}$, $Q_{\mathbf{\ddot{q}}}$, $Q_{\mathbf{T}}$, and $R_{\mathbf{u}}$ represent the positive definite matrices for stage cost and input cost respectively. 
$\mathbf{q}_0$ is the configuration of joint angles while standing and the first term of \eqref{eq:cost-function-design} minimizes the error between it and the final joint configuration of the optimization.
$\sum_{i=1}^{2} P_i T_i$ minimizes the travel time for Phases 1 and 2 with $P_i$ being positive scalars.
Next, $\delta$ is the relaxation variable for leg contact during Phase 2 and it is minimized by the cost term $\sum_{i=1}^{N_2} P_{\delta} \delta$.
All the costs above together minimizes the joint velocities and acceleration for all states to generate a smoother optimized trajectory with appropriate control inputs.

\subsection{Physical Constraints}
\label{subsec:physical-constraints}
The following constraints are added to the optimization to account for geometric limitations and physical hardware limitations:
\begin{itemize}
    \item Joint Angle: $\mathbf{q}_{min} \leq \mathbf{q}(t_k) \leq \mathbf{q}_{max}$
    \item Joint Velocity: $\dot{\mathbf{q}}_{min} \leq \dot{\mathbf{q}}(t_k) \leq \dot{\mathbf{q}}_{max}$
    \item Torque: $\mathbf{u}_{min} \leq \mathbf{u}(t_k) \leq \mathbf{u}_{max}$
    \item Friction Cone: $ \abs{\frac{T_{Fx}(t_k)}{T_{Fz}(t_k)}}, \abs{\frac{T_{Bx}(t_k)}{T_{Bz}(t_k)}}  \leq \mu$
    \item Contact Force:  $ T_{z, min} \leq T_{Bz}(t_k), T_{Fz}(t_k) \leq T_{z, max}$
    \item Contact Position: $-\delta \leq z_{C} \leq 0$
\end{itemize}
Here, the joint position, velocity, and torque limitations are considered under physical hardware constraints~\cite{bosworth2015super} and 
the friction coefficient $\mu$ for the contact legs is assumed as a value of 0.5.
The rear feet contact phase is the most significant part for the jumping performance, where we enforce contact force and position constraints.
The maximum contact force $T_{z, max}$ is estimated with physical hardware limitation and a minimum contact force $T_{z, min}$ is included to ensure the optimized force is large enough to support the weight of the robot.
$z_{C}$ represents the vertical position of the contact point of the back leg on the ground, which should be zero but it is relaxed with a variable $\delta$ to enhance the feasibility of the local planner. 

\subsection{Safety Constraints for Obstacle Avoidance}
\label{subsec:obstacle-avoidance}
\begin{figure}
    \centering
    \includegraphics[width=\linewidth]{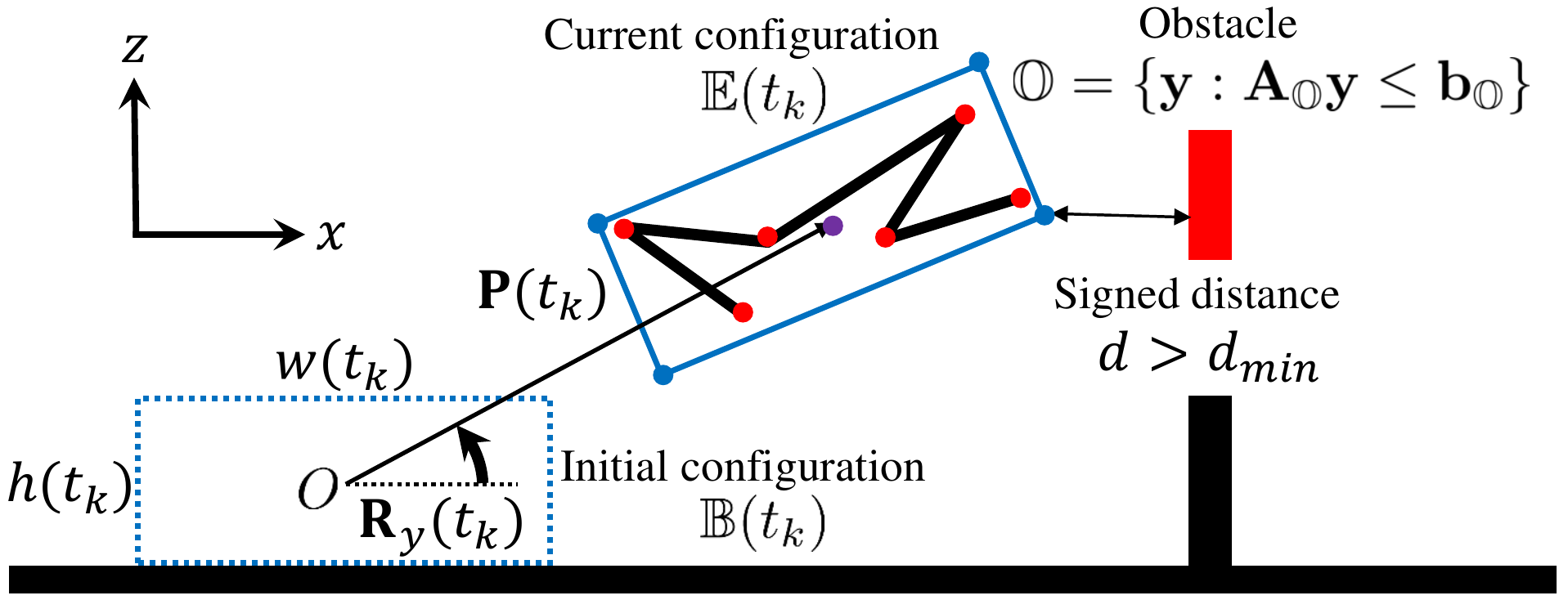}
    \caption{Obstacle avoidance between quadruped in flight phase and window-shaped obstacle.
    The key points of the robot to construct bounding box are marked as blue and the center of the bounding box is marked as red.
    Safety constraint between the bounding box and the convex obstacle (colored in red) is maintained by forcing the signed distance $d$ between them to be larger than a safety margin $d_{min}$.}
    \label{fig:obstacle-avoidance}
\end{figure}
Obstacle avoidance provides an additional constraint for the optimization in which a convex set that represents the bounding-box of the robot maintains a minimum distance from convex obstacles \cite{ zeng2020differential,zhang2018autonomous}, shown in Fig. \ref{fig:obstacle-avoidance}.
The quadruped is modeled as a controlled object $\mathbb{E}(t_k)$ with a rectangular bounding-box encapsulating the model throughout the trajectory.
We model the bounding-box as a convex set by utilizing six key points $\mathbf{p}_i (t_k)$ (the red points in Fig. \ref{fig:obstacle-avoidance}) on the robot: two points on hip, knee, and feet, respectively, whose coordinates are denoted as $\mathbf{p}_i = [x_{p_i}, z_{p_i}]^T \in \mathbb{R}^2$.
The center of the bounding box $\mathbf{p}_c(t_k)$ (the purple point in Fig. \ref{fig:obstacle-avoidance}) is determined by the mean position of the key points,
\begin{equation}
    \mathbf{p}_c(t_k) = \dfrac{1}{6} \sum_{i=1}^6 \mathbf{p}_i (t_k)
\end{equation}
and the rotation $\mathbf{R}_{y}(t_k) = \mathbf{R}_{y, q_{\theta}} $ is assumed as the robot body pitch angle.
The controlled bounding box (the solid blue rectangle in Fig. \ref{fig:obstacle-avoidance}) could be regarded as a convex region $\mathbb{E}(t)$ transformed from an initial convex set $\mathbb{B}(t)$ with rotation $\mathbf{R}_{y}(t_k) \in \mathbb{R}^{2 \times 2}$ and translation $\mathbf{P}(t_k) \in \mathbb{R}^{2}$ at each time step.
\begin{equation}
    \mathbb{E}(t_k) = \mathbf{R}_{y}(t_k) \mathbb{B}(t_k) + \mathbf{P}(t_k).
\end{equation}
%
%
Here, $\mathbf{P}(t_k) = \mathbf{p}_c(t_k)$ and the initial bounding box (the dashed blue rectangle in Fig.~\ref{fig:obstacle-avoidance}) could be considered as a rectangle centered at origin $O$ with time-varying width $w(t_k)$ and height $h(t_k)$, which could be represented as $\mathbb{B}(t_k) = \{y : \mathbf{L}(w(t_k),h(t_k)) \mathbf{y} \leq \mathbf{l}\}$ where $\mathbf{L}\in \mathbb{R}^{4 \times 2}$. 

To ensure our bounding box is valid, we add an additional cost term as follow into \eqref{eq:cost-function-design} to minimize height and width along the trajectory,
\begin{equation}
    J_{w,h}(t_k) = \sum_{k} w(t_k)^2 + h(t_k)^2, \label{eq:cost-bounding-box}
\end{equation}
while enforcing the $i$-th key points to lie inside the four vertices $\mathbf{V}_j(t_k) \in \mathbb{R}^{2}$ (the blue points in Fig. \ref{fig:obstacle-avoidance}) of the bounding box with the following equality constraints
\begin{equation}
    \label{eq:keypoint-constraint}
    \mathbf{p}_i(t_k) = \sum_{j=1}^{4} \mathbf{V}_j(t_k) \gamma_{ij}(t_k).
\end{equation}
Here, $\gamma_{ij}(t_k)$ denotes the weight for $j$-th vertex under the convex representation of $i$-th key points at each time step $t_k$.
\begin{equation}
    \sum_{j = 1}^{4} \gamma_{ij}(t_k) = 1, \gamma_{ij}(t_k) \geq 0 
\end{equation}

When the robot intends to jump through the obstacle, the upper and lower boundaries of a window-shaped obstacle are considered as a combination of two obstacles. Each of them can be described as a convex set $\mathbb{O}$ with a rectangular shape, 
\begin{equation}
    \mathbb{O} = \{\mathbf{y} \in \mathbb{R}^2 : \mathbf{A}_{\mathbb{O}} \mathbf{y} \leq \mathbf{b}_{\mathbb{O}}\}.
\end{equation}
The obstacle avoidance optimization algorithm is equivalent to a dual optimization problem at each time step $t_k$ as follows, with the minimum signed distance $d$ being larger than a safety margin $d_{min}$, 
\begin{equation} \label{eq:obstacle-avoidance-constraint}
    \begin{aligned}
        -\mathbf{l}^{T}\mu(t_k) + (\mathbf{A}_{\mathbb{O}} \mathbf{P}(t_k) - \mathbf{b}_{\mathbb{O}})^{T}\lambda(t_k) &> d_{min} \\
        \mathbf{L}^T \mathbf{\mu}(t_k) + \mathbf{R}_{y}(t_k)^T \mathbf{A}_{\mathbb{O}}^T \mathbf{\lambda}(t_k) &= 0 \\
        ||\mathbf{A}_{\mathbb{O}}^T \mathbf{\lambda}(t_k)|| < 1, \; \mathbf{\lambda}(t_k) > 0 , \; \mathbf{\mu}(t_k) &> 0.\\
    \end{aligned}
\end{equation}
Here, $\mathbf{\lambda}$ and $\mathbf{\mu}$ are dual variables related to the obstacle $\mathbb{O}$. The detailed proof of \eqref{eq:obstacle-avoidance-constraint} with duality can be found in~\cite{zhang2020optimization}.
The additional cost in \eqref{eq:cost-bounding-box} together with key points constraints in \eqref{eq:keypoint-constraint} and obstacle avoidance constraints in \eqref{eq:obstacle-avoidance-constraint} allows the robot to change its intrinsic dimensions to jump through a height-constrained obstacle.

Additionally, the final state constraint is proposed as follow,
\begin{equation}
    q_x(t_{N+1}) \geq x_{obs},
\end{equation}
which ensures that the robot lands past the obstacle in its final configuration, where $x_{obs}$ represents the $x$ position of the obstacle.
This constraint is necessary, otherwise, the local planner might generate a safe but conservative trajectory without jumping past the obstacle. 

%% file: sections/planners.tex
\section{Planners}
\label{sec:planners}
In order to utilize the optimized jumping maneuver to travel through a congested space, the jumping mode is embedded into an autonomous navigation framework with localization and mapping, motion planners, and a gait decision maker as introduced in this section. 

This framework works as follows. Two maps are generated and updated in this pipeline. One is a 2D occupancy grid map encoding the traversable region for the robot, and a 2.5D map keeps track of the obstacle height in the environment. 
A global planner searches for a global path on the occupancy map and sends an in-range target waypoint to the state machine.
Based on the coordinate of waypoint, the state machine decodes the  obstacle height from 2.5D height map and decides whether the robot should use walking or jumping modes. If a change of modes is detected, the state machine also adds a standing mode to smooth the mode transitions.  
Moreover, the state machine also adds desired yaw rotation based on the selected modes.
The target waypoint and desired yaw position are tracked by a local planner when the walking mode is chosen.

\begin{figure}
    \centering
    \includegraphics[width=\linewidth]{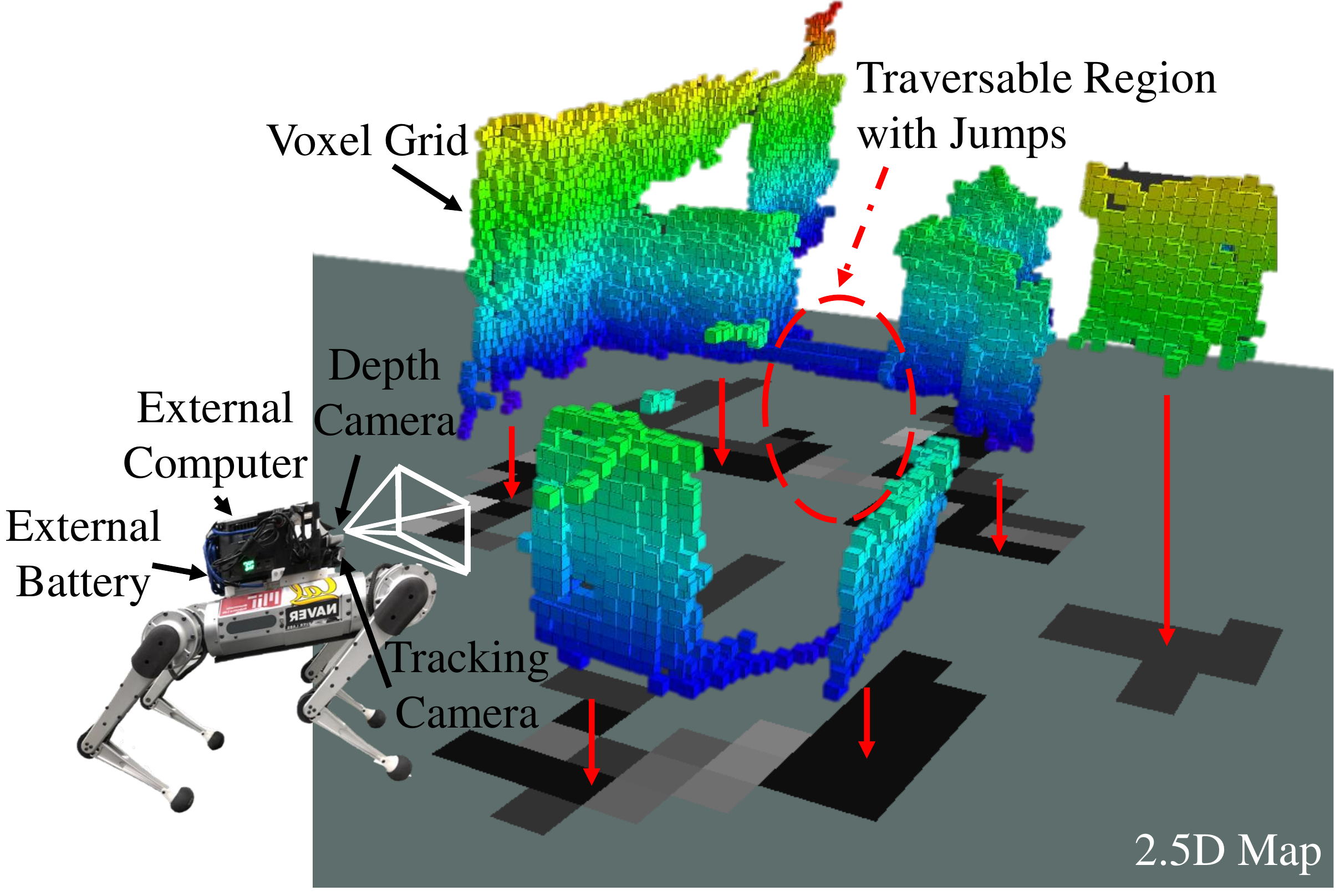}
    \caption{The Mini Cheetah carries a sensor suite that contains an RGB-Depth camera, a tracking camera, and an additional power source and computer for online navigation. The robot uses the tracking camera to estimate its current state. The 3D environment represented by a voxel grid is perceived by the RGB-Depth camera. A 2.5D map that records the obstacle height in each grid is updated from the voxel grid. The darkness of the grid grows with increasing obstacle height. A region that has obstacles whose heights are below the robot's jumps are classified as a traversable region with jumps.}
    \label{fig:mapping}
\end{figure}

\subsection{Mapping and Localization}
Additional onboard computing and sensors are included to enable the robot to navigate through the environment. 
A sensor suite, as shown in Fig. \ref{fig:mapping}, includes an Intel NUC providing external computing resource with a RGB-Depth camera (Intel RealSense D435i), and a tracking camera (Intel RealSense T265).  The additional computer and sensors are used to accurately map and localize the robot while exploring the environment. 

RTAB-Map is used to generate a 3D point cloud of heights in the environment \cite{labbe2019rtab}. 
From the voxel grid with a resolution of $0.1$~m, two maps are created, an occupancy map with a resolution of $0.1$~m and a 2.5D map with a resolution of $0.2$~m.
The maximum height of the voxel grid of the same planar coordinates is recorded.
If this height is above what the robot can jump over, this grid is marked as untraversable with a Boolean variable $0$, otherwise, the grid will be a free space encoded by Boolean variable $1$, and this information is stored in a 2D occupancy map.
A coarser 2.5D map is also created along with the voxel grid, which features a map of discrete tiles sizes, \textit{i.e.}, $0.2$~m. 
The maximum height within the tile is then marked as obstacle height $z^{obs}$ in that tile, as illustrated in Fig.~\ref{fig:mapping}. 

The observed robot current states in the 3D space, $[\tilde{q}_{x,y,z,\phi}, \tilde{\dot{q}}_{x,y,z,\phi}]^T$, are estimated by Visual Inertial Odometry~(VIO) through the tracking camera. 

\subsection{Global Planner}
To find a global path from the current robot position to the given goal location, a search-based A* planner over the updated 2D occupancy map is utilized to find a collision-free path that contains a sequence of 2D coordinates of waypoints. 
The global planner also considers an inflation of $0.4$~m of the occupied grid as a planning safety buffer. 
In the planned global path, a waypoint that is ahead of the robot's current position coordinate in the given range is selected. 
In this paper, we choose a waypoint that is $0.3$~m ahead of the robot. 
The 2D coordinate $[q^{des}_x, q^{des}_y]$ and its obstacle height $z^{obs}$ are recorded in the 2.5D map and sent to the following state machine and local planner.

\subsection{State Machine}
After receiving a target waypoint with the information of admissible height from the global planner, a state machine is developed to choose robot locomotion modes to jump, stand, or walk. 
Moreover, the state machine also decides a target yaw position of the waypoint. 
If the obstacle height is $0$~m, which indicates that it is a free space that the robot can walk through, the state machine sets the target yaw position as the direction between robot current coordinate and waypoint, \textit{i.e.}, $q^{des}_{\phi} = <[\tilde{q}_x, \tilde{q}_y]^T, [q^{des}_x, q^{des}_y]^T>$.
If the obstacle height is non-zero, the state machine will proceed to prepare the robot to jump. 
Firstly, the desired yaw is set to be perpendicular to the obstacle to jump over, and the position of the target waypoint will be held and not updated. 
This will lead the robot to reach to the obstacle and get prepared to jump. 
Once the robot arrives at the target coordinate, the state machine will firstly switch the robot from the walking mode to a standing mode, and then switch to the jump mode.
After the robot executes the jump and lands, the state machine will switch back to walking mode through a standing transition, and the target waypoint will be updated by the global planner.
Note that the obstacle height from the given waypoint is always admissible for the robot to jump over as the global planner will avoid obstacles that are beyond the robot's jumps ability in the 2D occupancy map. 

The output from the state machine contains discrete locomotion modes: jumping, standing, and walking, and target waypoint with desired yaw $q^{des}_{\phi}$.

\subsection{Local Planner for Walking}\label{subsec:local_planner}
If the state machine decides to use the walking mode of the quadrupedal robot to walk around the obstacles, a local planner that is able to track the given waypoint with orientation $[q^{des}_x, q^{des}_y, q^{des}_\phi]^T$ is designed.
This local planner leverages a collocation-based trajectory optimization using a simple double integrator dynamics to track the given waypoint. 
The optimization problem with collocation has $30$ nodes and spans over $1$ second.  
The decision variables include the states of robot planar and yaw positions and their derivatives. 
It is formulated to minimize the distance between the final node to the target waypoint while respecting the robot's actual velocity limits.
The optimized trajectory contains a sequence of robot desired walking velocities, $\dot{q}^{des}_x$, $\dot{q}^{des}_y$, and turning rates $\dot{q}^{des}_{\phi}$, and will be tracked by the robot using a walking controller.

\begin{figure*}[!htp]
    \centering %
    \begin{subfigure}{\textwidth/4}
      \includegraphics[width=0.98\linewidth]{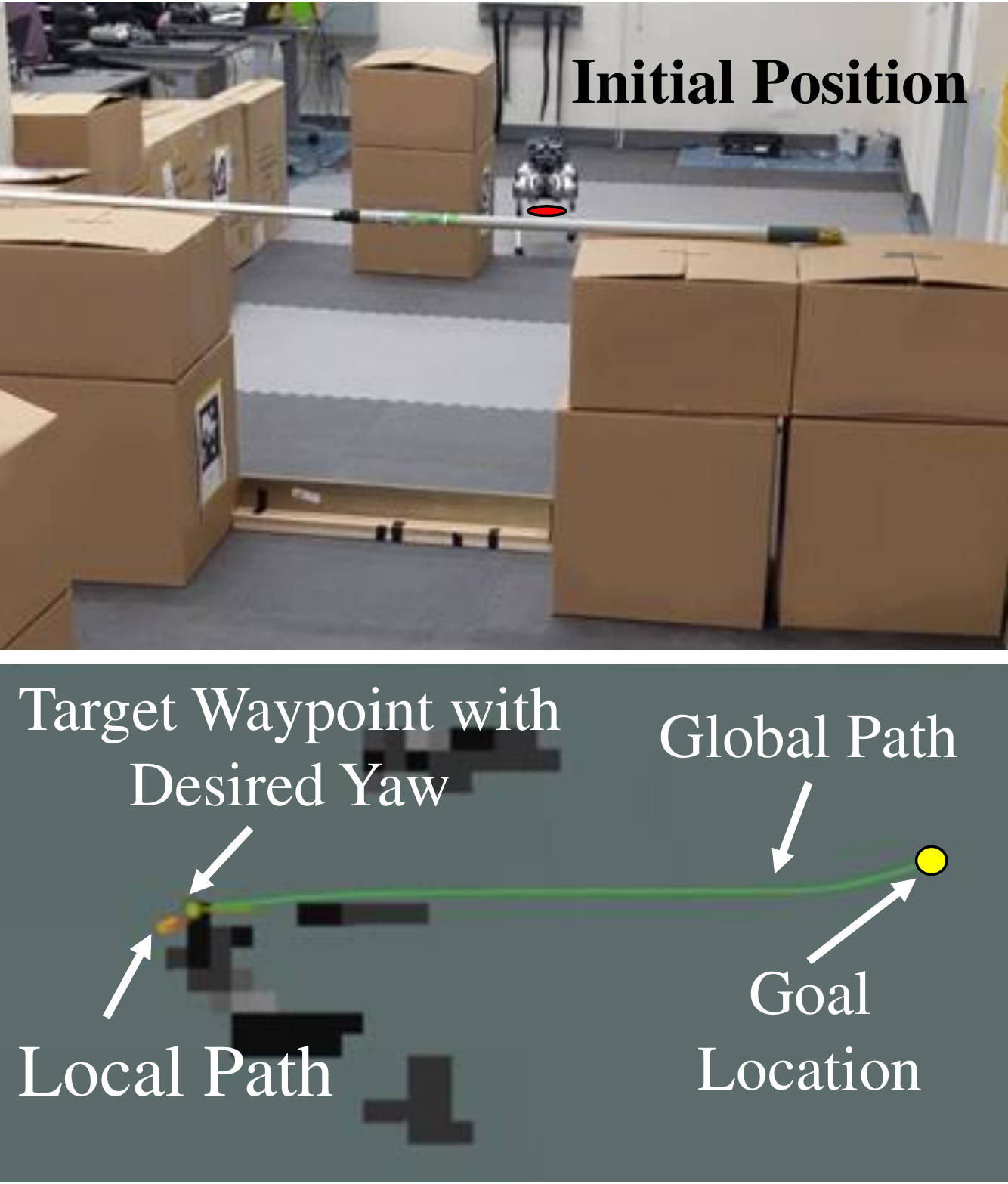}
      \caption{}
    \label{fig:snapshota}
    \end{subfigure}\hfil %
    \begin{subfigure}{\textwidth/4}
      \includegraphics[width=0.98\linewidth]{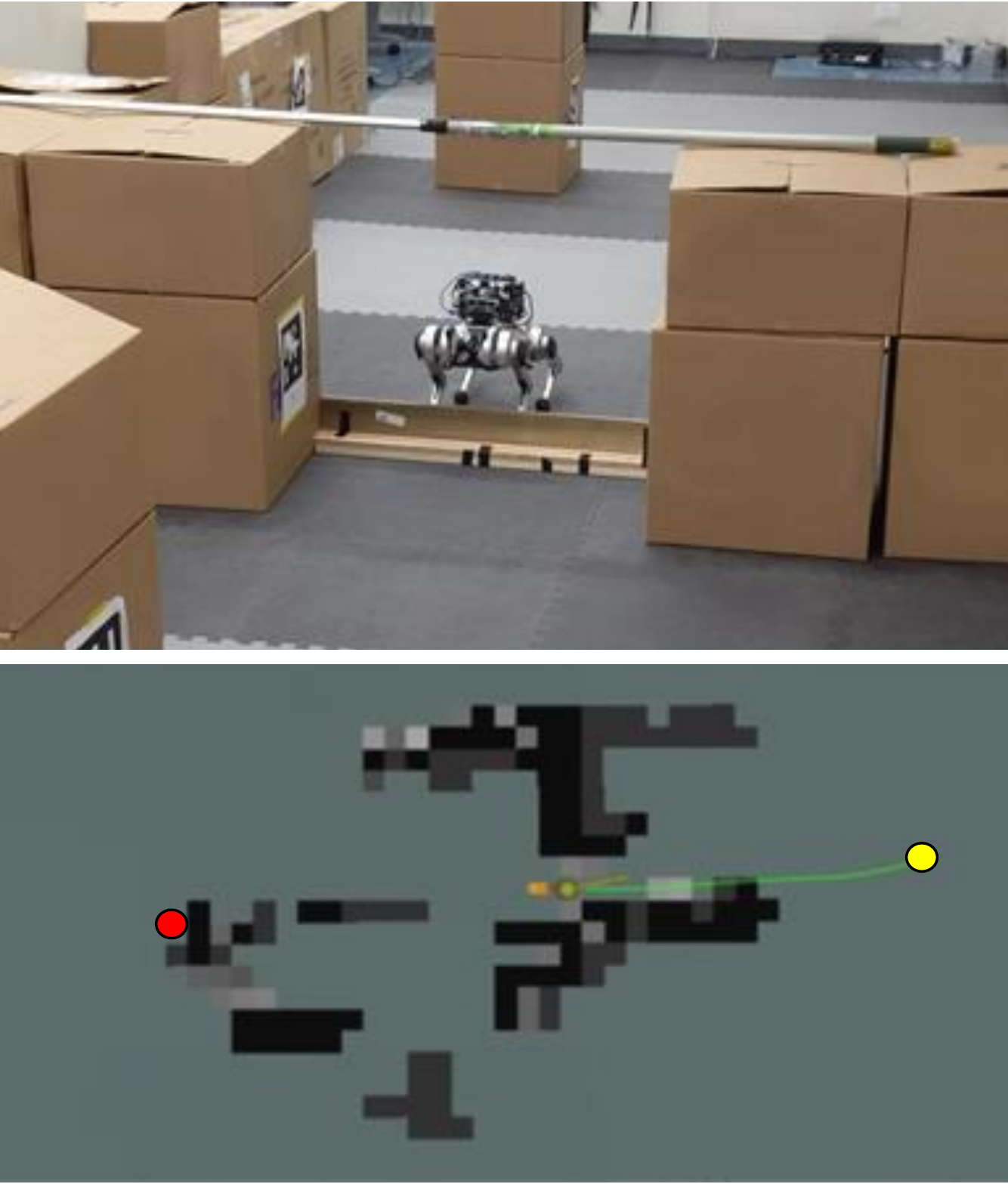}
      \caption{}
    \label{fig:snapshotb}
    \end{subfigure}\hfil %
    \begin{subfigure}{\textwidth/4}
      \includegraphics[width=0.98\linewidth]{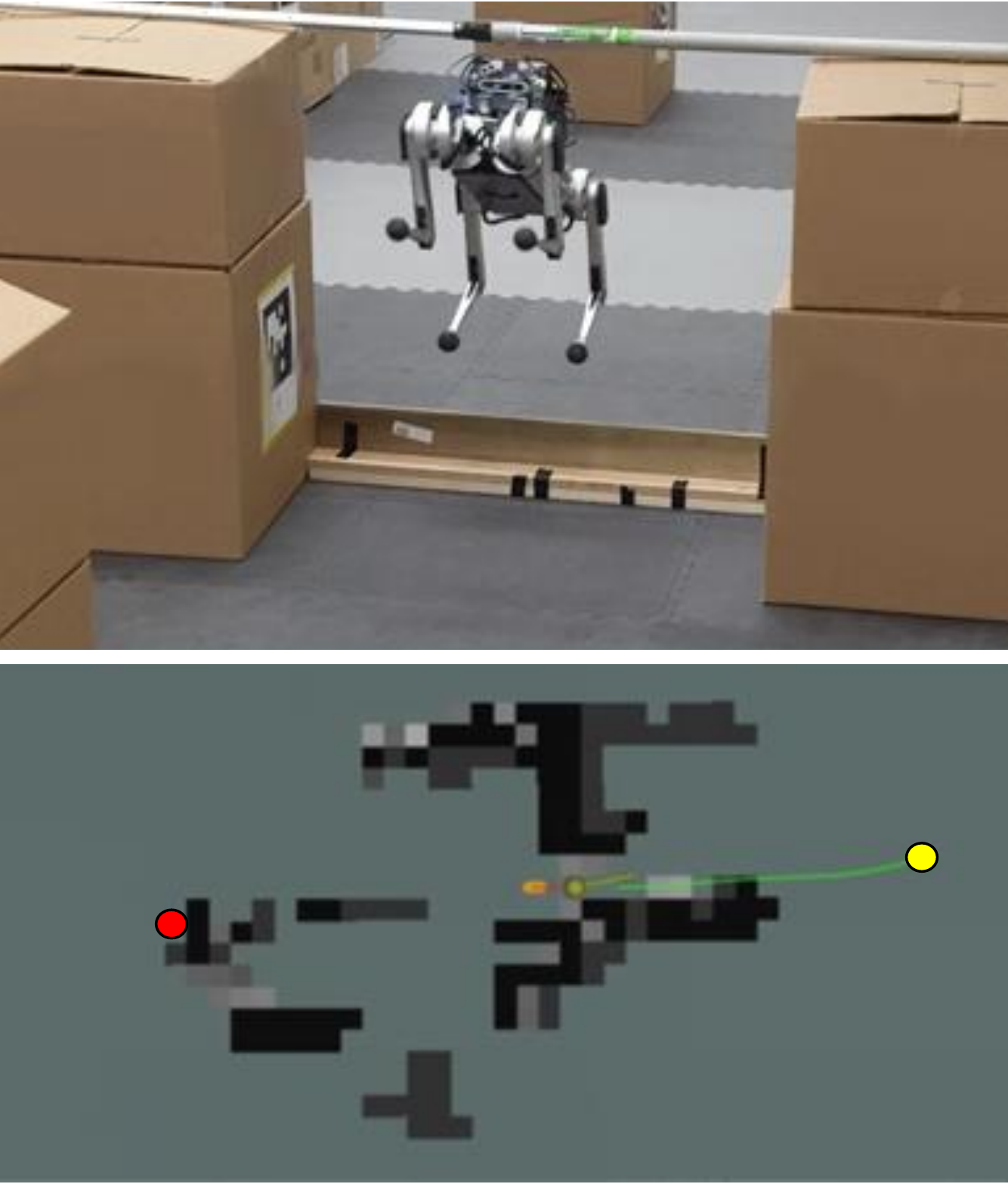}
      \caption{}
    \label{fig:snapshotc}
    \end{subfigure}\hfil %
    \begin{subfigure}{\textwidth/4}
      \includegraphics[width=0.98\linewidth]{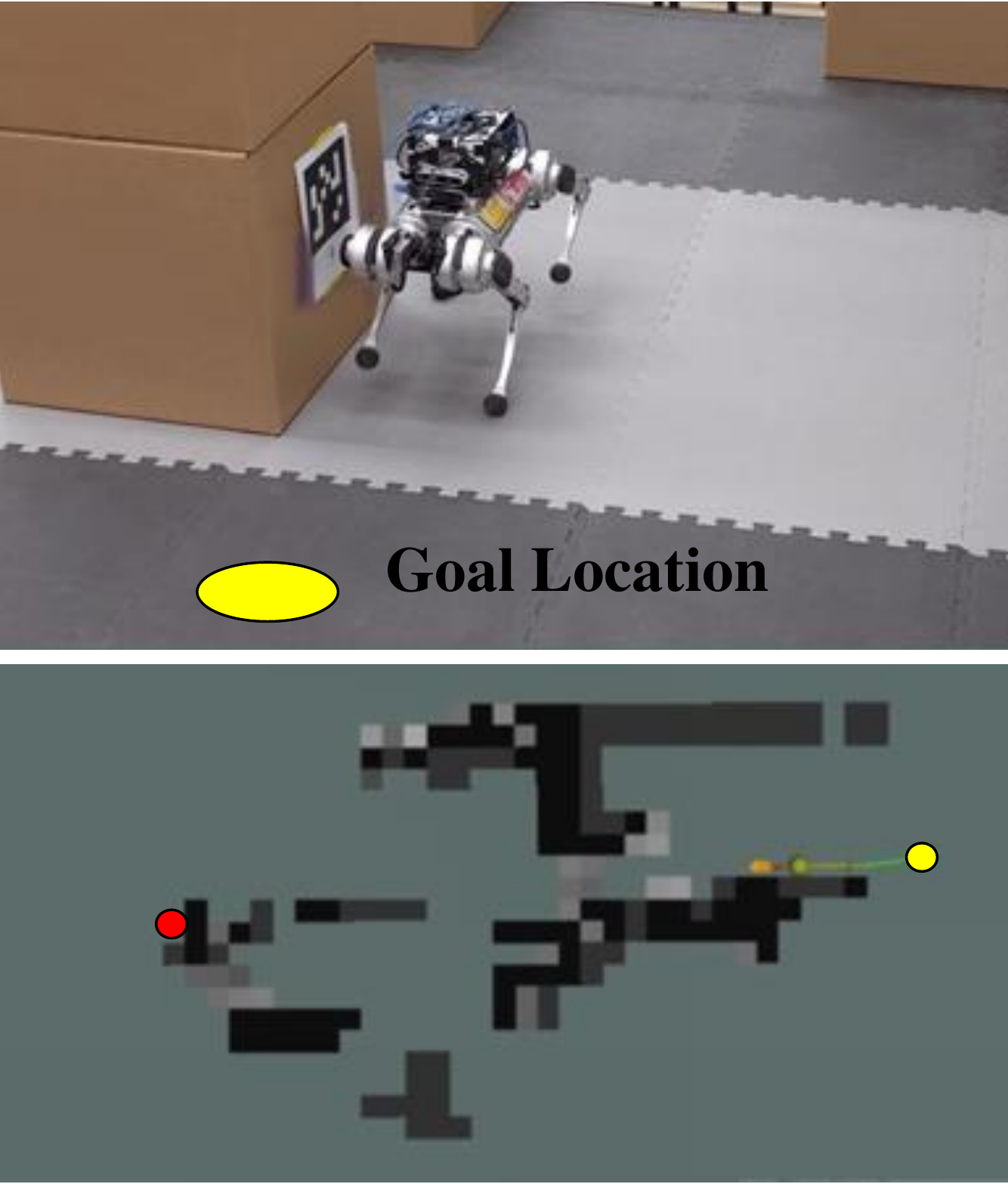}
      \caption{}
    \label{fig:snapshotd}
    \end{subfigure}\hfil %

    \caption{A snapshot of the Mini Cheetah: (a) avoiding an obstacle within the environment while navigating to the end goal; (b) preparing to jump over an obstacle that obstructs the robot from the goal; (c) robot in flight jumping through the obstacle; and (d) continuation of navigating the environment to the end goal. Top-view of the mapped environment in RViz is shown in the bottom plots. The robot initial position and goal location are marked as red and yellow points, respectively. The global path is represented by green line while the local path is denoted as yellow points. The target way point with desired yaw orientation given from the state machine is shown as green point.}
    
    \label{fig:snapshot}
\end{figure*}

\begin{figure*}[!htp]
    \centering %
    \begin{subfigure}{\textwidth/3}
      \includegraphics[width=\linewidth]{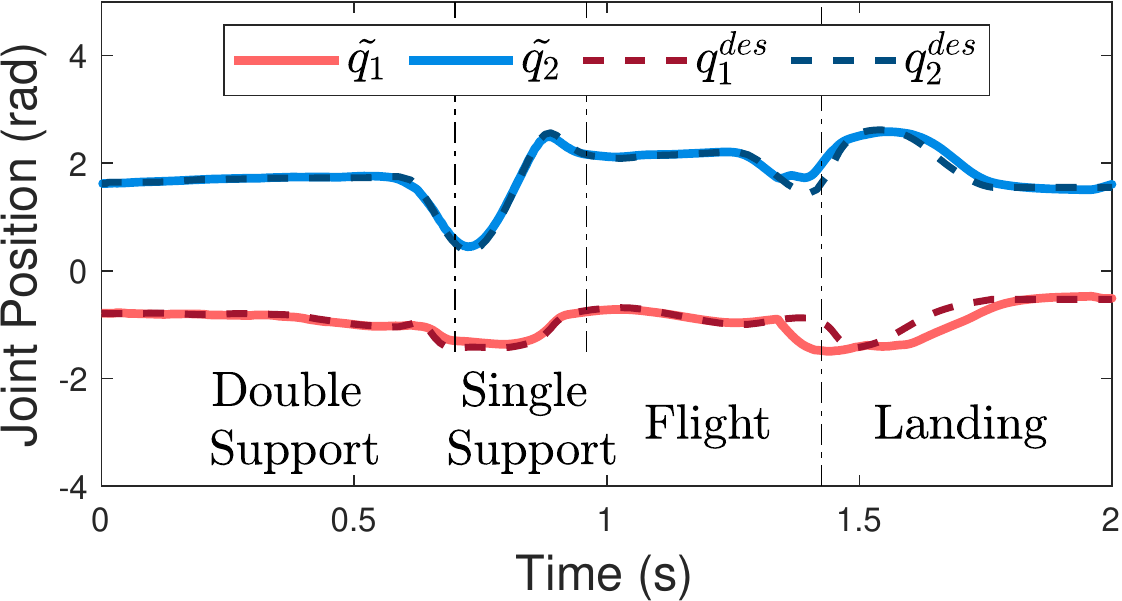}
      \caption{Front Leg Joint Angles}
    \label{fig:plota}
    \end{subfigure}\hfil %
    \begin{subfigure}{\linewidth/3}
      \includegraphics[width=\linewidth]{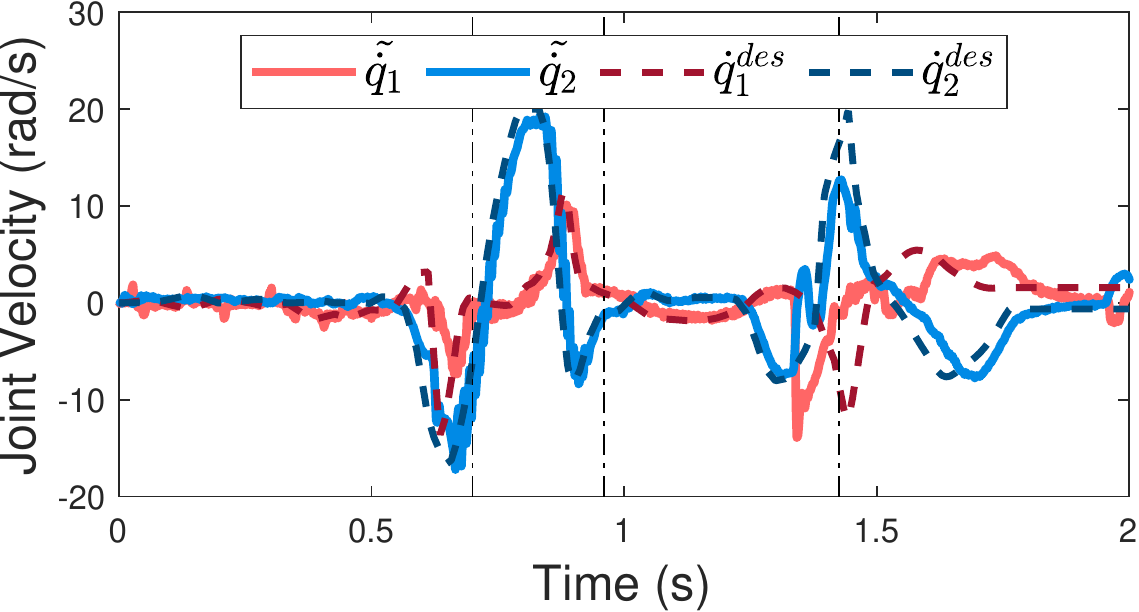}
      \caption{Front Leg Joint Velocities}
    \label{fig:plotb}
    \end{subfigure}\hfil %
    \begin{subfigure}{\linewidth/3}
      \includegraphics[width=\linewidth]{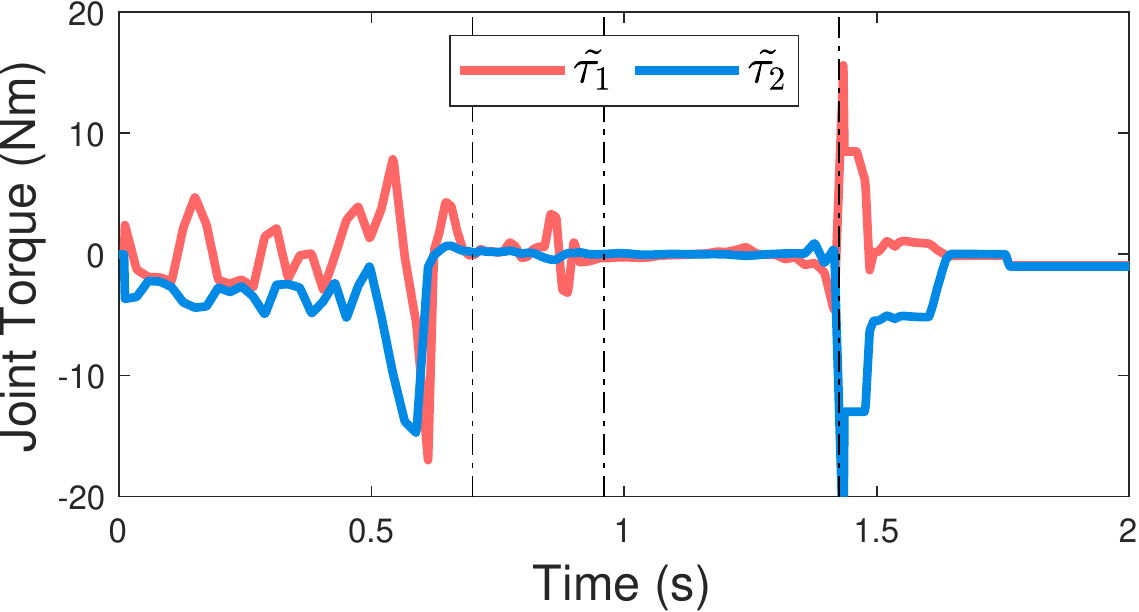}
      \caption{Front Leg Joint Torques}
    \label{fig:plotc}
    \end{subfigure}\hfil %
    
    \begin{subfigure}{\textwidth/3}
      \includegraphics[width=\linewidth]{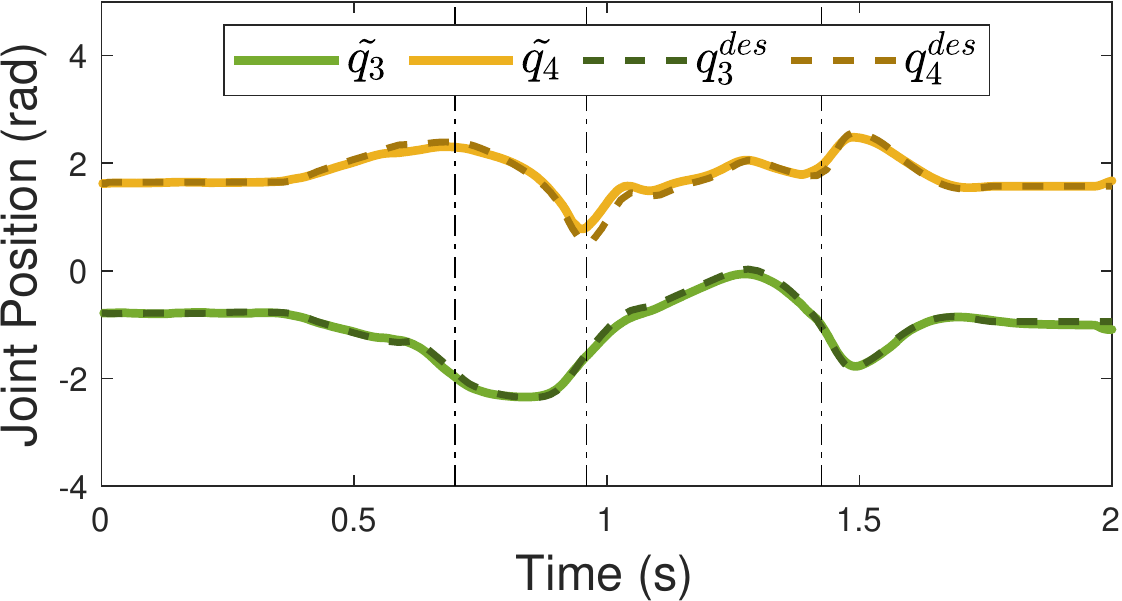}
      \caption{Back Leg Joint Angles}
    \label{fig:plotd}
    \end{subfigure}\hfil %
    \begin{subfigure}{\linewidth/3}
      \includegraphics[width=\linewidth]{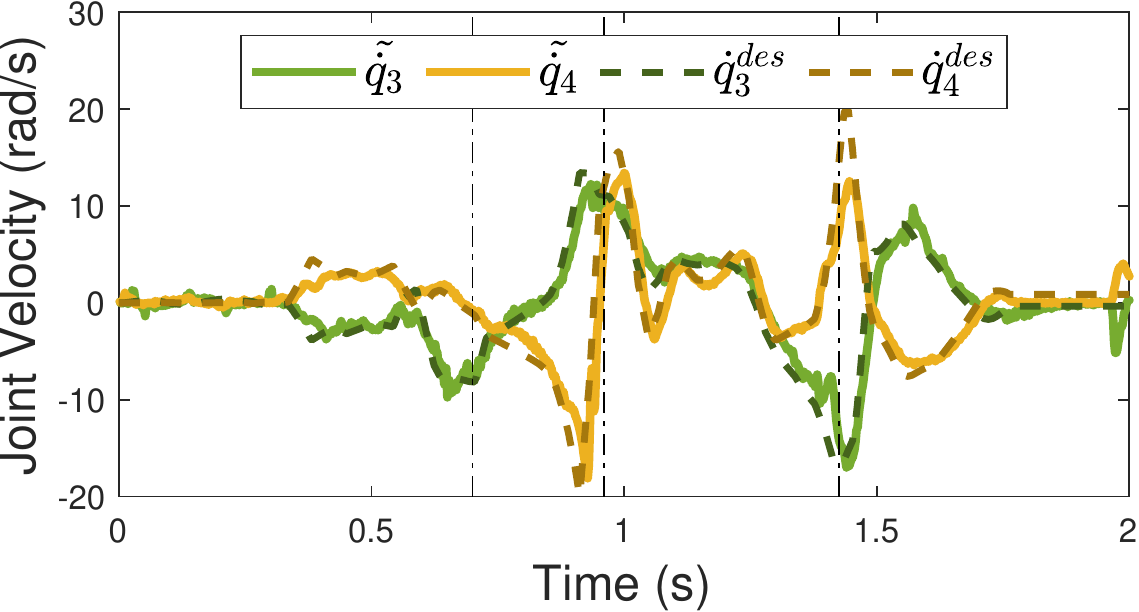}
      \caption{Back Leg Joint Velocities}
    \label{fig:plote}
    \end{subfigure}\hfil %
    \begin{subfigure}{\linewidth/3}
      \includegraphics[width=\linewidth]{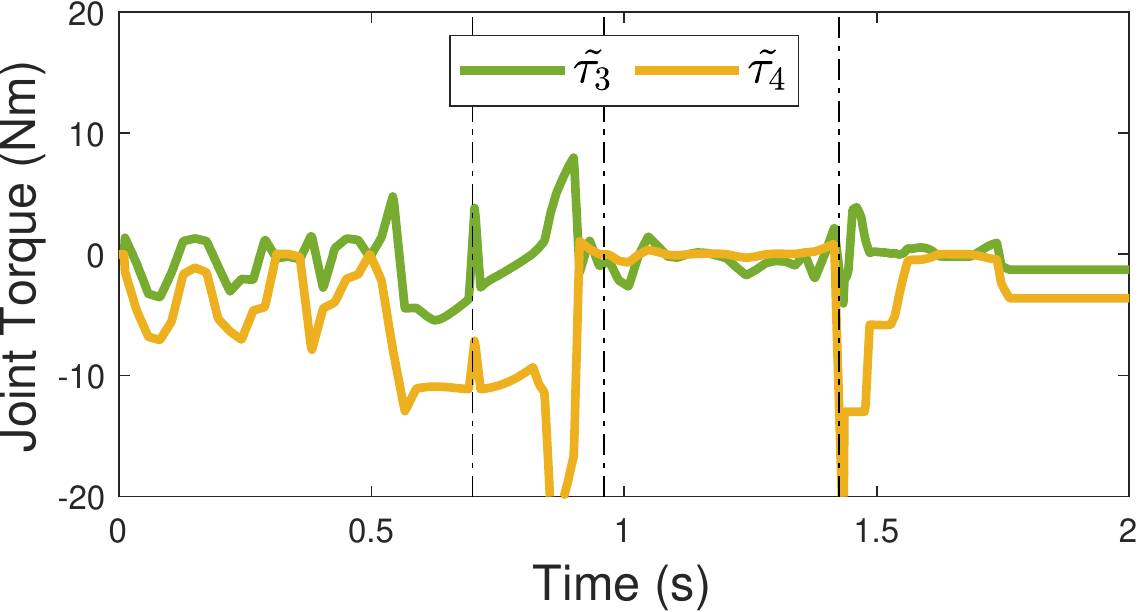}
      \caption{Back Leg Joint Torques}
    \label{fig:plotf}
    \end{subfigure}\hfil %

    \caption{The comparison between the measured joint angles, velocities, and torques from the experiment, which are shown in solid lines, and the desired values from the jumping trajectory optimization drawn in dash lines. The values plotted are gathered from the hip and knee joints of the front right and back right legs of the robot. The front hip joint is shown as red, the front knee joint is demonstrated as blue, back hip is drawn as green and the back knee is marked as yellow.}
    
    \label{fig:plots}
\end{figure*}

%% file: sections/controller.tex
\section{Control}
\label{sec:control}
After having seen trajectory generation and planners for two locomotion modes, we are going to present the controllers that are used to enable quadrupedal robots to realize these locomotion behaviors, where jumping and walking controllers are described in Sec. \ref{subsec:jumping-control} and \ref{subsec:walking-control}, respectively.  

\subsection{Jumping Control}
\label{subsec:jumping-control}
For a robot jump, the trajectory optimization in Sec.~\ref{sec:trajectory-generation} solves for a reference trajectory with the desired joint angles $\mathbf{q}^{des} \in \mathbb{R}^{4 \times N}$, joint velocities $\mathbf{\dot{q}}^{des} \in \mathbb{R}^{4 \times N}$, and torque inputs $\bm{\tau}^{des} \in \mathbb{R}^{4 \times N}$.
The reference joint trajectory and torque input are then linearly interpolated with a 1 kHz sample rate 
and are sent to a joint-level PD controller with a feed-forward term of desired torque, as follow, 
\begin{equation} \label{eq:joint-pd}
    \bm{\tau} = \bm{\tau}^{des} + \mathbf{K}_{p,i}(\mathbf{q}^{des} - \mathbf{q}) +  \mathbf{K}_{d,i}(\mathbf{\dot{q}}^{des} - \mathbf{\dot{q}})
\end{equation}
where $\mathbf{K}_{p,i} \in \mathbb{R}^{3 \times 3}$ and $\mathbf{K}_{d,i} \in \mathbb{R}^{3 \times 3}$ are proportional and derivative gains during the $i$-th phase and $\bm{\tau}^{des}$ is the feed-forward torque obtained from the optimization for jumping. 

Gain scheduling of both $\mathbf{K}_{p,i}$ and $\mathbf{K}_{d,i}$ occur so that the gains for each phase of the jump can be tuned independently. 
During the landing phase, the proportional and derivative gains are interpolated which respect the landing phase's duration, \textit{i.e.}, $\bm{\tau}^{des}_{landing} = \lambda \bm{\tau}^{des}_{landing}$, where the feed-forward torque is decreased by a factor, $\lambda$, to reduce the impact of timing mismatch has when the quadruped re-establishes contact with the ground.
By manipulating both the gains and feed-forward torque during landing, the quadrupedal robot is more robust to errors in the landing time, where the optimal landing trajectory is used. 
By interpolating the gains, the spike of motor torque upon landing will be reduced while still allowing the robot to reach the desired end configuration. 
This results in safe landing maneuvers even in the scenarios where the robot's landing configuration deviates from the planned one.

\subsection{Walking Control}
\label{subsec:walking-control}
A model predictive controller (MPC) and whole body impulse controller (WBIC) was developed in~\cite{kim2019highly} for the Mini Cheetah. 
Empowered by this controller, the robot is able to track given planar velocity and yaw rotation rate from the local planner developed in Sec.~\ref{subsec:local_planner} while maintaining gait stability.

%% file: sections/results.tex
\section{Experiments and Evaluation}
\label{sec:results}
Using the approach described throughout this paper, the performance was evaluated based upon the Mini Cheetah's ability to navigate through environments, to detect and jump over the obstacle, and to continue navigating to the goal location. 
This approach is validated with experiments using different environments, \textit{i.e.}, obstacle configurations, and is demonstrated in the supplementary video\footnote{\url{https://youtu.be/5pzJ8U7YyGc}}.
The environment contains obstacles of various sizes obstructing the path of the Mini Cheetah, shown in Fig.~\ref{fig:snapshot}, and some of them cannot be traversed only by walking.
Without having the ability to jump over certain obstacles, an environment with large obstructions would prevent quadrupeds from reaching the goal location. 

The Mini Cheetah with the carried sensor suite is able to detect obstacles and create maps of the environment. 
While navigating through the environment, the Mini Cheetah accurately marks higher obstacles and safely walks around them, shown in Fig.~\ref{fig:snapshota}. 
Once an obstacle that can be jumped over is found, the Mini Cheetah tunes its orientation to point towards the obstacle and navigates in front of it. 
Afterward, the Mini Cheetah can successfully perform the jumping maneuver and clear the obstacle.
Fig.~\ref{fig:snapshotb} demonstrates the Mini Cheetah overcoming a $13$ cm tall obstacle with a window size of $70$ cm that would normally not be traversable by walking. 

Fig.~\ref{fig:plots} shows the tracking performance of the joint angles and velocity for all four phases of the jump over a $13$ cm tall obstacle. 
The PD controller running at 1 kHz provides great tracking of both the joint position and velocity throughout the duration of the jump while obeying the physical hardware limitations. 
In Fig.~\ref{fig:plots}, spikes in joint velocity indicate the transition between all feet contact and rear feet contact at 0.7 s in Fig.~\ref{fig:plotb}, transition from rear feet contact to flight at 0.9 s in Fig.~\ref{fig:plote}, and the impact of landing at 1.4 s in Fig.~\ref{fig:plotb}, \ref{fig:plote}.
In the optimization-based trajectory generation, the landing phase is optimized with a lower coefficient of friction than the take off phase to prevent the possibility of over rotation of the robot upon landing with a high horizontal velocity. 
To reduce the overall tendency to bounce upon landing, gains are interpolated throughout the landing phase. 


Experiments throughout this paper are nearing the maximum jumping heights due to the hardware limitations of the Mini Cheetah. 
With the current torque to weight ratio of the Mini Cheetah, the robot is able to clear obstacles up to 13 cm tall with the 2.25 kg sensor suite attached. 
Additional experimentation is conducted without the sensor suite, and the Mini Cheetah is capable of clearing obstacles up to 24 cm.
Both experiments conducted are able to demonstrate that the optimization problem can solve for optimal trajectories with different torque to weight ratios.
Quadrupedal robots with larger torque to weight ratios using the methodology presented in this paper would therefore be capable of clearing higher obstacles than shown in experiments. 

%% file: sections/conclusion.tex
\section{Conclusion and future works}
\label{sec:conclusion}
In this paper, we introduced a navigation autonomy for quadrupedal robots to maneuver in cluttered environments while avoiding obstacles. We developed an end-to-end framework that enabled multi-modal transitions between walking and jumping skills. 
Using multi-phased collocation based nonlinear optimization, optimal trajectories were generated for the quadrupedal robot while avoiding obstacles and allowing the robot to jump through window-shaped obstacles. 
An integrated state machine, path planner, and jumping and walking controllers enabled the Mini-Cheetah to jump over obstacles and navigate previously nontraversable areas.


Future work includes implementing an online trajectory optimization and optimizing for jumps where the take off and landing are not at the same elevation.